# Influence of Pedestrian Collision Warning Systems on Driver Behavior – A Driving Simulator Study

Snehanshu Banerjee[1], Mansoureh Jeihani[2], Nashid K Khadem[2], Md. Muhib Kabir[2],

[1]* – Omni Strategy LLC, Baltimore MD
[2]* – Department of Transportation and Infrastructure Studies – Morgan State University, Baltimore, MD 21251
email: snban1@morgan.edu; nakha3@morgan.edu; mdkab1@morgan.edu; mansoureh.jeihani@morgan.edu

**Abstract**

With the advent of connected and automated vehicle (CAV) technology, there is an increasing need to evaluate driver behavior while using such technology. In this first of a kind study, a pedestrian collision warning (PCW) system using CAV technology, was introduced in a driving simulator environment, to evaluate driver braking behavior, in the presence of a jaywalking pedestrian. A total of 93 participants from diverse socio-economic backgrounds were recruited for this study, for which a virtual network of downtown Baltimore was created. An eye tracking device was also used to observe distractions and head movements. A Log logistic accelerated failure time (AFT) distribution model was used for this analysis, to calculate speed reduction times; time from the moment the pedestrian becomes visible, to the point where a minimum speed was reached, to allow the pedestrian to pass. The presence of the PCW system significantly impacted the speed reduction time and deceleration rate, as it increased the former and reduced the latter, which proves the effectiveness of this system in providing an effective driving maneuver, by drastically reducing speed. A jerk analysis is conducted to analyze the suddenness of braking and acceleration. Gaze analysis showed that the system was able to attract the attention of the drivers, as the majority of the drivers noticed the displayed warning. The familiarity of the driver with the route and connected vehicles reduces the speed reduction time; gender also can have a significant impact as males tend to have longer speed reduction time, i.e. more time to comfortably brake and allow the pedestrian to pass.

*Keywords – Connected vehicles, Pedestrian collision warning, Driving simulator, Eye tracking*

## 1. Introduction

Pedestrian safety is one of the main priorities in designing any traffic network. However, this category of road users is often referred to as Vulnerable Road Users (VRUs) due to their vulnerability to road accidents and fatalities [1, 2]. According to the National Highway Traffic Safety Administration (NHTSA), there were 5,977 pedestrian fatalities across the US in 2017 [3]. Moreover, pedestrian fatalities in urban areas increased by 46% from 2008 [3] and most of these fatalities resulted from driver distraction, obstruction of visibility or distraction and unawareness of an oncoming vehicle from the pedestrians perspective [4].

As a result, improving pedestrian safety has become the main goal of several transportation agencies, as they focus on developing warning systems to alert drivers, when pedestrians are on their path. The earliest of these systems were based on cameras and laser sensors [4]. Nevertheless, these systems were not that effective in improving the safety of pedestrians as these sensors have a fixed line-of-sight and angle of view which, in some instances, are not capable of detecting pedestrians behind certain obstacles or around building corners, as well as, in bad weather conditions [4, 5]. To overcome these drawbacks, new systems that integrate pedestrians into the vehicles' network through smartphones, known as vehicle-to-pedestrian (V2P) and pedestrian-to-vehicle (P2V) systems, or more generally, vehicle-to-everything (V2X), has been developed [6]. The main advantage of these systems is that drivers will be able to know the presence of pedestrians even when they are obstructed by an object or in the dark [7]. In general, V2P and P2V systems aim to communicate between the vehicles and pedestrians via a wireless technology and cellular networks and send alerts, either to the driver or the pedestrians, about the presence of the other party via text messages and app alerts [2].

Hence, several research studies have been conducted in order to test the effectiveness of the V2P and P2V warning systems in enhancing the safety for the road users through assessing different driving experience parameters, with and without the use of these systems. These research studies can be classified under two main categories: those that examined the system's performance and those that examined the driving and/or pedestrian behavior and performance after using the V2P or P2V systems; albeit the vast majority belongs to the first category. In addition, there were two testing techniques used in these research studies, to test the performance of the pedestrian warning systems which were either through driving simulators or field tests.

As for the pedestrian warning systems' safety performance, Hussein et al. [8] evaluated the collision prediction for a V2P and P2V smartphone application and, through a field test, the system was able to detect the approaching vehicle, five seconds earlier than the in-vehicle sensing devices. Wang et al. [9] used a field experiment to test the car detection performance and maximum detection distance of a V2P application called "Walksafe". Through this experiment, the system was able to correctly detect the approaching cars in 77% of the cases (total of 109 cases), with only three cases of false positive detection, at a maximum detection distance of 50 meters, were recorded. Similarly, Li et al. [10] assessed the precision ratio, using the potential arrival area (PAA) measure, of a V2P system called V2PSense. In the field experiment, the researchers designed six routes, each requiring eight minutes to traverse, and calculated the PAA, every minute. Through this experiment, it was found that the precision ratio for all six routes was above 85%, ranging from 86.9% to 98.8%, with an average of 92.6%. While Liu et al. [11] assessed the safety performance of a Pedestrian-Oriented Forewarning System (POFS) in terms of the probability of alerting the pedestrian at the right time. The researchers conducted a field experiment at 30km/hr and 60km/hr and found that the probability is at least 90% at both speeds. He et al. [2], found, through a field experiment, that the use of a V2P system will decrease the collision risk significantly, even reaching zero. Agreeing with these results, Sugimoto, Nakamura and Hashimoto [12] conducted a field experiment to assess the performance of a P2V system and found that the collision risk between the vehicle and pedestrians is greatly reduced as the driver can stop the vehicle at 5m before reaching the pedestrian.

On the other hand, some studies examined the driving behavior and performance after using the V2P or P2V systems. Of these studies, Kim et al. [13] assessed the drivers' behavior and performance when using a pedestrian warning system through a field experiment using distance-based measures. The researchers used eye-tracking glasses and in-vehicle cameras, as well as, GPS data to measure the change in driver's behavior after receiving warnings from the system. Two of the parameters used in this study to evaluate the drivers' behavior; are gaze on pedestrian, and foot-on brake pedal. Regarding the gaze on pedestrian, distances decreased after presenting the warning in the far pedestrian condition as compared to no warning condition by 19.31% in the monoscopic display and 25.21% in the volumetric display. As for the foot-on-brake distances, when drivers were presented with warnings, this parameter decreased by 32.44% and 36.15% in the monoscopic and volumetric conditions in the near pedestrian condition, respectively. Similarly, Rahimian et al. [7] tested the behavior of pedestrians with a smartphone which has a V2P application that detected the motion of vehicles and pedestrians and warned the pedestrians with an auditory alarm about the incoming vehicle, but did not provide instructions of whether to cross the road or not. Through measuring the gaze direction of the pedestrians, using a specially fitted glasses, this experiment assessed the number of collisions that occurred as well as the pedestrians' attention to traffic. By analyzing the data, there were 10 collisions out of the 28 trials; however, a warning was sent for each of these collisions; while, the pedestrians who received warnings spent less amount of time looking at traffic compared to those who did not receive it. However, Rahimian et al. [14] did not find a significant difference between pedestrians who received alerts and those who did not, in terms of amount of time spent before crossing. In addition, Cummings, Huang and Clamann [15] designed and tested a smartphone application to warn pedestrians of incoming vehicles before crossing the street. In this field experiment, the researchers tested the pedestrians' behavior to the alerts received in terms of their decision to perform unsafe or risky crossings. The results showed that of those who received late alerts, there were 16 unsafe crossings, or 33%, and of those who received the alerts just in time, there were 161 risky crossings, or 20%.

Connected and Automated vehicle (CAV) technology is still in testing phases throughout the world while people have to pay higher premiums to get automated vehicle (AV) features, when they buy a new car, and as such, majority of the common people do not have the opportunity to drive a car with such safety features. Although a lot of researches have been conducted to evaluate influence of V2P systems on pedestrian reaction times, these studies were limited and had a lot of constraints, including driving at specific speeds [12, 16]. The authors also found no studies that evaluate driver braking behavior involving a P2V warning system, using a driving simulator. In this study, a virtual network with P2V technology was created, where people were invited to drive with no speed constraints, and their driving behavior and experience, when subjected to this CAV technology, were evaluated.

## 2. Methods

*2.1. Driving simulator*

A medium-fidelity full scale driving simulator at the Morgan State University Safety and Behavioral Analysis (SABA) Center and a Tobii Pro eye tracking device [17] was used for this study, to analyze driver behavior in response to a pedestrian collision warning (PCW), using P2V technology. The VR-Design studio software developed by FORUM8 Co. [18] was used to develop a virtual network of downtown Baltimore in Maryland. The idea behind choosing this location was such that, majority of the participants being familiar with the downtown Baltimore area, the virtual network creates a more realistic driving experience for them. The simulator has the capability capture data such as steering wheel control, braking, acceleration and speed while the eye tracking device can capture head movements, eye gaze and gaze durations, all in real time. The driving simulator and the eye tracking glasses are shown in Figure 1.

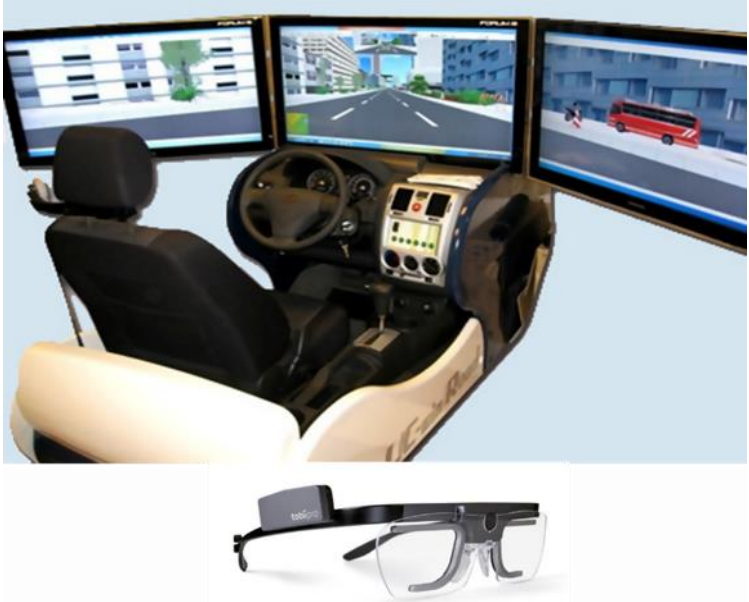

**Figure 1. Driving simulator at Morgan State University and Tobii Pro eye tracking glasses**

*2.2. Scenario and network design*

A network was built replicating downtown Baltimore, in the Inner Harbor area. Two scenarios were designed for this study, one with the PCW and the other without the warning i.e. the base scenario. The base scenario was tested first, followed by some other scenarios involving different CAV features and then the scenario with PCW, to avoid bias or the learning effect of the driving simulator. A major one-way four-lane road, Pratt Street, with a speed limit of 30 mph, in downtown Baltimore, was designed using the VR-Studio software. Pratt street has a lot of foot traffic and as such would be an ideal location to evaluate a PCW system, when encountering a jaywalking pedestrian. A level of service B, i.e. light traffic was used in these scenarios, so that the participants do not slow down, due to high traffic which may have been the case otherwise, creating issues evaluating the PCW system. Pratt street is a complete street, with a 14-foot-wide wide shared bus and bike lane, three 12-foot lanes and wider sidewalks. The proposed PCW system takes advantage of the P2V technology on the pedestrian's smartphone to broadcast the device's position to the warning enabled vehicle. For both the baseline and PCW system scenarios, as soon as the participant crosses a waypoint, a jaywalking pedestrian appears at an approximate distance of 40 meters from the waypoint, which then broadcasts the device/pedestrian location to the oncoming vehicle. The distance of 40 meters was chosen based on visibility, traffic conditions, road geometry and most importantly NHTSA guidelines on stopping distance, to avoid collision. According to the guidelines, to avoid a collision, the initial gap between the vehicle and the pedestrian, should be greater than the stopping sight distance of the vehicle. It is expressed as [19]:

$$D_0 > \frac{-V_{vi}^2}{2a_v} \qquad (1)$$

where, $D_0$ is the initial gap between the vehicle and the pedestrian, $V_{vi}$ is the initial speed and $a_v$ is the acceleration/deceleration of the vehicle. A snapshot for this event is shown in Figure 2.

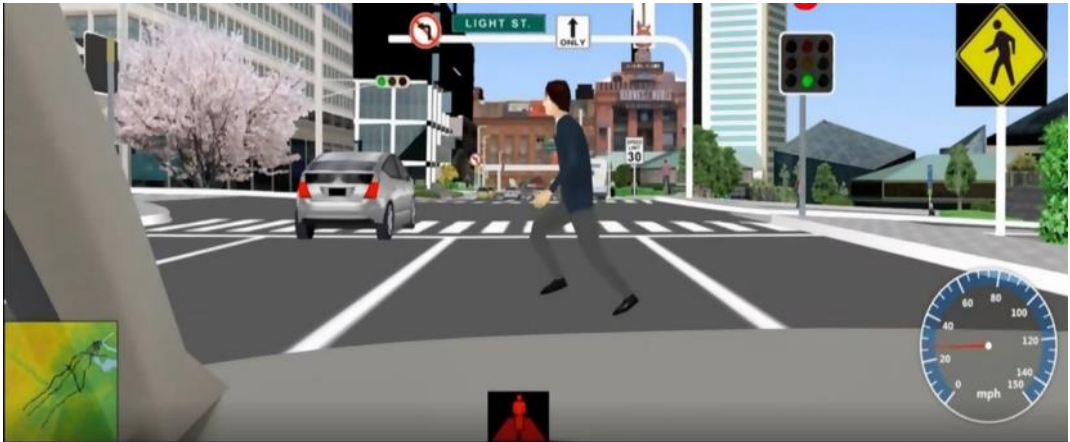

**Figure 2. A snapshot of the driving simulator environment**

*2.3. Participants*

Institutional Review Board (IRB) approval was received before human participants were recruited, where 93 participants from diverse socio-economic backgrounds took part in this study. Participants were recruited through a combination of emails to participants from prior studies involving the driving simulator [20-27], distribution of flyers across the university and throughout Baltimore County. Participants were paid at the rate of $15 per hour of driving, for their involvement in the study. The study was briefly explained to the participants and they were given an opportunity to get familiar with the driving simulator. Two pre and post scenario survey questionnaires were developed to gather information on sociodemographics, CAV experiences and experience using the driving simulator. Table 1 presents some of the descriptive statistics of the participants as well as selected pre driving questionnaires.

**Table 1. Participant socio-demographics**

| Variables | Characteristics | Percentage |
|---|---|---|
| Gender | Female | 44 |
|  | Male | 56 |
| Age | 18-25 | 37 |
|  | 26-35 | 29 |
|  | 36-45 | 14 |
|  | 46-55 | 12 |
|  | >55 | 8 |
| Education Level | High School or less | 12 |
|  | College degree | 61 |
|  | Post-graduate | 27 |
| Household income level | <$20,000 | 27 |
| | $20,000 - $49,999 | 34 |
| | $50,000 - $99,999 | 22 |
| | >$100,000 | 17 |
| Familiarity with CAVs | AVs only | 27 |
| | CVs only | 5 |
| | CAVs | 24 |
| | None | 43 |
| Familiarity with downtown Baltimore | Yes | 59 |
| | Somewhat | 34 |
| | No | 7 |

*2.4. Hazard-based duration model*

Hazard-based duration models are probabilistic methods that are used to evaluate cases that have a definite origin point until the occurrence of an event [28]. In the transportation field, these models have been used to study a number of time-related events such as assessing critical factors impacting crash durations and thus developing crash duration prediction models [29-31], evaluating the impacts of cellphone usage on driver reaction time in response to a pedestrian crossing the road [32, 33], modeling the duration of highway traffic incidents [34-36], etc. The duration variable used in this study, is the speed reduction time, which is calculated from the moment, the jaywalking pedestrian becomes visible to the participant driving the simulator, until a minimum speed is reached i.e. the participants lets the pedestrian cross or comes to a complete stop. A proportional hazard and an accelerated failure time (AFT) model, are two of the approaches that could have been used for this analysis. These models are used to evaluate the influence of covariates on the hazard function. As compared to the hazard model where the hazard ratios are assumed to be constant over time, the AFT model enables the covariates to accelerate time in a survivor function, when all covariates are zero, resulting in easier interpretation [37]. Based on this, an AFT modeling approach was selected for this study.

*2.5. Longitudinal jerk analysis*

Jerk is termed as the rate of change in acceleration or deceleration. Jerk can be either positive or negative. Positive longitudinal jerk values are associated with the abrupt release of the brakes, immediately followed by pressing the gas or throttle while negative longitudinal jerk values are associated with the abrupt release of the gas or throttle, immediately followed by pressing the brakes. The magnitudes of both the positive and negative jerks depend on the abruptness and extent to which the brakes and the throttle are pressed or released.

The boundaries for comfortable jerk were found to be $\pm 1 m/s^3$ while the acceptable limits of longitudinal jerk were defined as $\pm 2 m/s^3$ [38]. Based on these limits, jerks can be classified as:

$$-1 \text{ m}/s^3 \leq \text{Jerk} \leq 1 \text{ m}/s^3 \rightarrow \text{Comfortable positive or negative jerk}$$
$$-2 \text{ m}/s^3 \leq \text{Jerk} \leq -1 \text{ m}/s^3 \rightarrow \text{Acceptable negative jerk}$$
$$1 \text{ m}/s^3 \leq \text{Jerk} \leq 2 \text{ m}/s^3 \rightarrow \text{Acceptable positive jerk}$$
$$\text{Jerk} < -2 \text{ m}/s^3 \text{ Or Jerk} > 2 \text{ m}/s^3 \rightarrow \text{Unsafe jerk values}$$

## 3. Results and Discussion

186 experiments were conducted; however, in 83 of those the participants either failed to yield to the pedestrian or missed them completely due to over speeding; thus, the final dataset used in the analysis consisted of 103 observations of the participants' braking maneuvers. The braking maneuvers of the participants, the moment they encounter the jaywalking pedestrian, were analyzed, in both the baseline scenario as well as the scenario involving a PCW system.

The average deceleration rate ($d_m$) at the moment the jaywalking pedestrian became visible until the participants slowed down to let the pedestrian pass or came to a complete stop is given by [39]:

$$d_m = \frac{V_i^2 - V_{min}^2}{2(L_{V_{min}} - L_{V_i})} \quad (2)$$

where: $V_i$ is participant's initial speed as they approach the waypoint where the jaywalking pedestrian first becomes visible, $V_{min}$ represents participant's minimum speed reached during the deceleration phase, $L_{V_i}$ is the distance between the vehicle's location when the initial speed was recorded at the waypoint and the point at which the jaywalking pedestrian starts crossing the street, and $L_{V_{min}}$ represents the distance between the vehicle's location when the minimum speed was recorded at the waypoint and the point at which the pedestrian starts crossing the street.

The speed reduction time ($S$) is calculated as the elapsed time between the participant's initial speed ($V_i$) and the minimum speed ($V_{min}$) reached to allow the pedestrian to pass, before accelerating.

A plot of the participants' speed profile before and after the waypoint, 40 meters from where they first possibly spotted the pedestrian, was generated. Through this plot as shown in Figure 3, several parameters related to the participants' braking maneuvers were calculated. It can be seen that participants braked harder, when they receive a PCW, compared to when not receiving a warning.

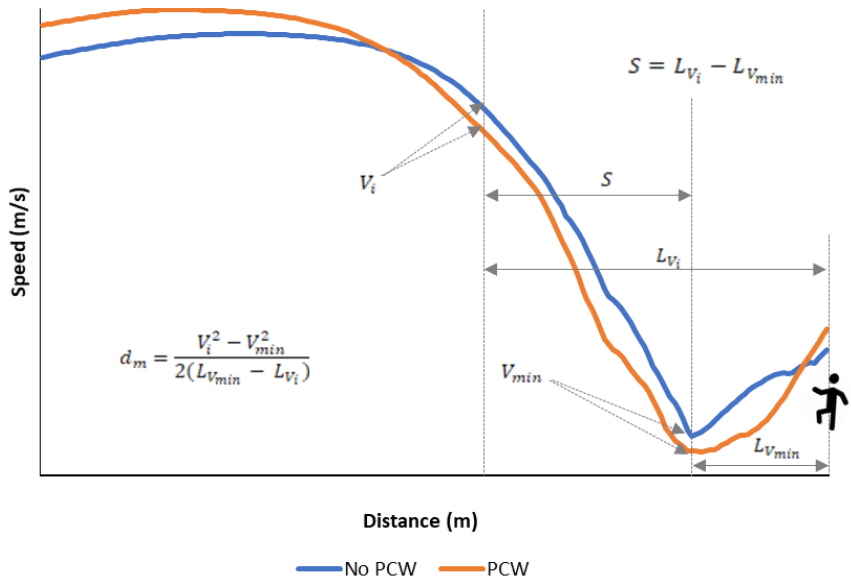

**Figure 3. Participant speed profile comparison**

To determine the braking behavior with and without the PCW, a plot of the average deceleration rates of all participants are plotted in Figure 4.

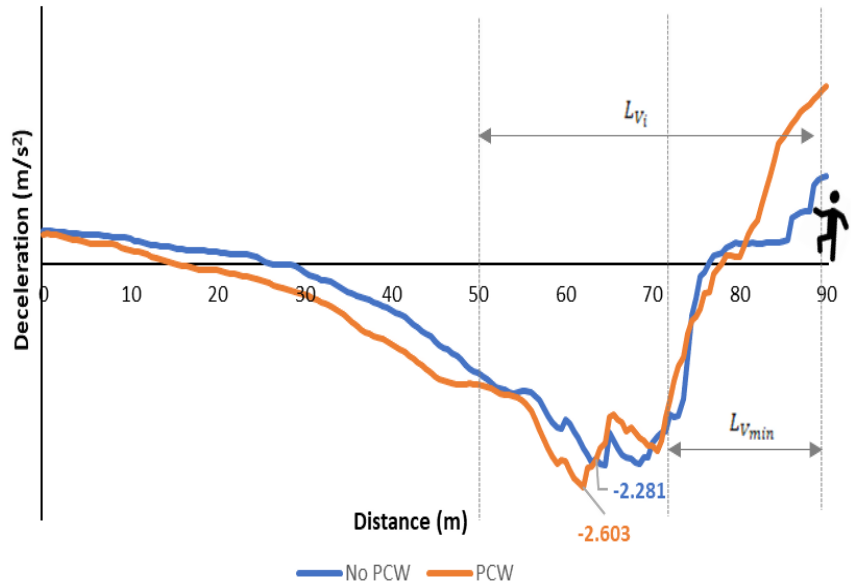

**Figure 4. Participant average deceleration**

The deceleration rate in Figure 4 shows that, the participants braked earlier, at the onset of the PCW. Average maximum deceleration (-2.603 m/s$^2$) for the PCW scenario is attained at 12 meters from the warning point, while maximum deceleration (-2.281 m/s$^2$) for the Non-PCW scenario is attained at 13.5 meters from the potential warning point. To confirm this braking behavior, the perception reaction time taken to release the throttle and the brake execution time from the moment the throttle is released until the initial brake application, is calculated for each participant. The braking intensity is calculated, one second after the brakes are pressed to determine the intensity, where on a scale of 0 to 1, 0 is no brake force and 1 is maximum brake force. The average reaction and braking execution time statistics are shown in Table 2.

**Table 2. Reaction and braking execution time statistics**

|  | Average Perception Reaction Time (s) | Average Brake Execution Time (s) | Average Time to Reach Max Deceleration (s) | Average Max Braking Intensity | Average Max Speed Change (m/s) |
|---|---|---|---|---|---|
| PCW | 0.29 | 0.2 | 1.75 | 0.56 | 5.53 |
| No PCW | 0.36 | 0.2 | 2 | 0.47 | 6.03 |

From Table 2, it can be seen that, participants react quicker in the presence of a PCW system. The average perception reaction time is quicker in the PCW scenario, as the participants get the warning before they can anticipate the pedestrian, and thus start braking early, gradually slowing down to let the pedestrian pass. The average time to reach maximum deceleration is 1.75 seconds and 2 seconds respectively, for the PCW and Non-PCW scenario which confirms the hard-braking behavior. The average maximum braking intensities at these points were 0.56 and 0.47 respectively. The average maximum speed change is the difference in speed from the warning point until the average maximum deceleration is reached. The slightly higher speed change (6.03 m/s compared to 5.53 m/s) can be attributed to the distance traversed in the additional 0.25 seconds.

From Table 2, it can be said that, participants took slightly longer to react in the absence of a PCW system. The braking intensity and the total duration of the perception reaction time, and brake execution time with a PCW system (total duration of 0.49 sec), show that hard braking is involved in the initial stages of a PCW system. The hard braking in the PCW scenario can be confirmed as maximum declaration is reached at 1.75 seconds compared to the 2 seconds, in the Non-PCW scenario, supported by the average maximum braking intensity (0.56 compared to 0.47) from Table 2, at the maximum deceleration points.

*3.1. Jerk analysis*

Jerk is calculated for each participant and averaged over every second, from the waypoint where the PCW is issued. The participants average jerk, every tenth of a second, is shown in Figure 5.

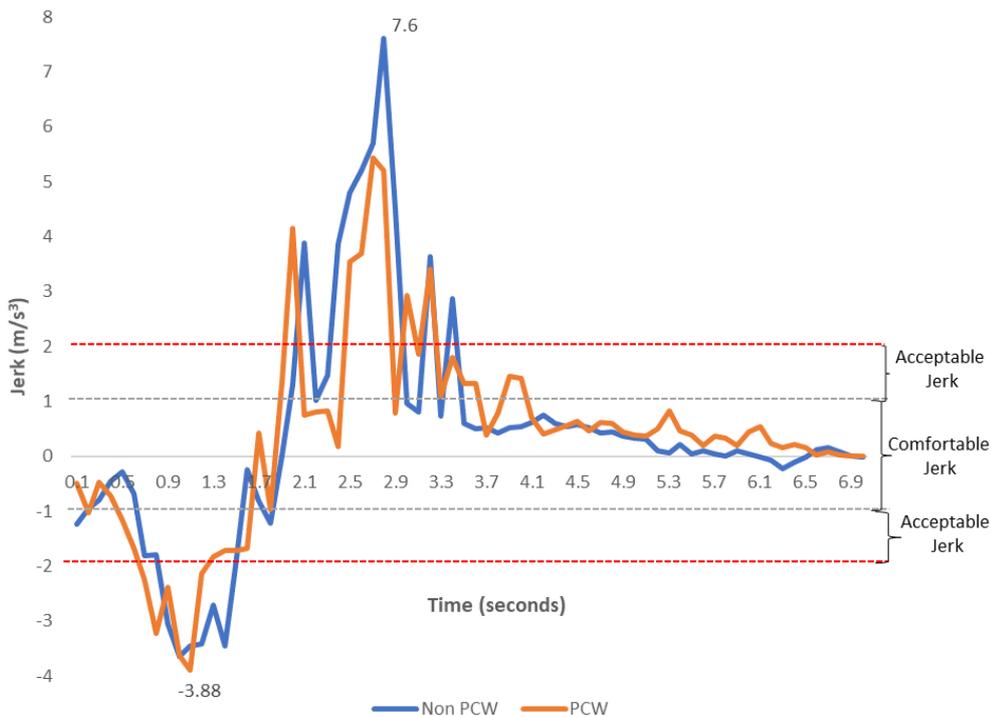

**Figure 5. Participants Average Jerk by Time**

It can be seen that, in the scenario with the PCW, the average initial jerk around 1.1 second after the warning is issued, is way below the acceptable jerk limit [38] and is thus unsafe. On the other hand, in the scenario without the PCW, around 2.8 seconds after the waypoint, the jerk is the maximum at 7.6 m/s$^3$. This being a positive jerk, it is from the sudden release of the brakes and can be termed as a highly uncomfortable jerk. The jerk drops sharply at 2.8 seconds and is still positive, which means that, the participants in the Non-PCW scenario, press the throttle suddenly. This signifies that the participants slowed down enough, to let the jaywalking pedestrian pass, before accelerating.

*3.2. Log logistic AFT model*

An ANOVA analysis revealed that there was a statistically significant difference in speed reduction times between the baseline scenario and the scenario with a PCW system. The speed reduction times were longer in the scenario involving the PCW system (mean difference in time = 0.61 seconds, $\rho \leq 0.05$) at 3.14 seconds compared to 2.53 seconds in the baseline scenario. This implies that the overall deceleration rate when the PCW was used, is less compared to when the system was not used (2.99 m/s$^2$ vs. 3.19 m/s$^2$). Although, this infers a smoother braking maneuver, it is not the case as seen in Figure 4. Since deceleration rate or braking behavior is affected by CAV technology, in this case a PCW system, a hazard-based duration model was developed to comprehend the participant's braking behavior in terms of speed reduction times. As demonstrated by Bella and Silvestri [39], this dependent variable is a positive duration dependence event as its probability increases as a result of an increase in the available time. A distribution assumption of the speed reduction time variable is required to estimate the hazard and the survival functions in a parametric setting. The hazard function gives the conditional failure rate while the survival function is the probability of a longer speed reduction time than a specified time. The most commonly used are the lognormal, log-logistic, exponential and Weibull distribution functions. In order to select the best fit and most applicable function, the Akaike information criterion (AIC), which is one of the most well-known approaches for model selection based on their adequacy [40, 41], and log-likelihood values were used. The four distributions were assessed, and it was found that the log-logistic model provides the best fit for the data as it had the lowest AIC values at -382.73 and the highest log-likelihood values at -128.158, among them. The hazard function $h(t)$ of the log-logistic duration model is expressed as [42]:

$$h(t) = \frac{\lambda p t^{p-1}}{1 + \lambda t^p} \tag{3}$$

with $p > 0$ and $\lambda > 0$ and the survival function $S(t)$ of the log-logistic duration model is expressed as [42]:

$$S(t) = \frac{1}{1 + \lambda t^p} \frac{1}{1 + (\lambda^{\frac{1}{p}} t)^p} \tag{4}$$

where $\lambda$ and $p$ are the location and the scale parameters, while $t$ is the specified time, respectively. Table 3 shows the descriptive statistics of the different parameters used in the log-logistic model and the speed reduction times of the participants in both the baseline scenario as well as the scenario involving the PCW system.

Table 3. Speed reduction time and Log logistic AFT variable descriptives

| Variables | Mean Value (No warning) | Std. Dev (No Warning) | Mean Value (PCW) | Std. Dev (PCW) |
|---|---|---|---|---|
| $V_i$ (m/s) | 11.18 | 3.11 | 10.90 | 3.33 |
| $L_{V_i}$ (m) | 50.67 | 0.44 | 50.55 | 0.34 |
| $V_{min}$ (m/s) | 1.85 | 1.92 | 1.33 | 1.76 |
| $L_{V_{min}}$ (m) | 70.63 | 9.16 | 71.77 | 7.98 |
| $d_m$ (m/s$^2$) | 3.19 | 1.04 | 2.99 | 1.45 |
| Speed Reduction Time(s) | 2.53 | 0.62 | 3.14 | 0.95 |

Moreover, in order to assess the goodness of fit for the log-logistic model, a plot of the cumulative hazard rate was determined from the model estimates and then utilized to build an empirical estimate of the cumulative hazard model. As seen in Figure 6, the points representing the estimate of the cumulative hazard function almost follow the 45° reference line; hence, it can be inferred that the participants' predicted speed reduction time, using the log-logistic model, is a good fit with the observed data.

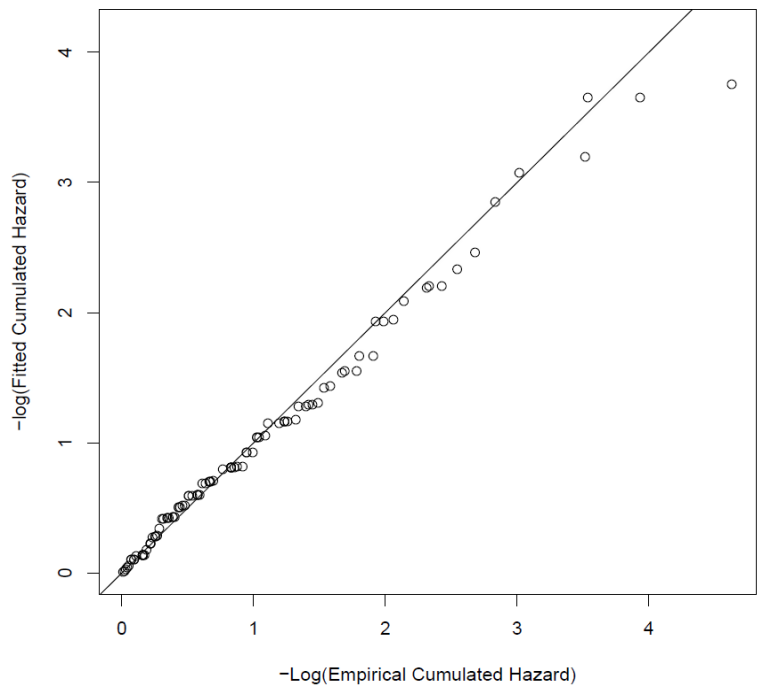

Figure 6. Cox-snell residuals for Log logistic AFT

The estimates from the Log logistic AFT model with the speed reduction times of the participants as the dependent variable are shown in Table 4.

**Table 4. Log logistic AFT parameter estimates**

| Variables | Estimate | Std. Error | z | ρ | Exp (β) |
|---|---|---|---|---|---|
| (Intercept) | 12.317 | 70.911 | 0.173 | 0.862 | 223521.809 |
| $V_i$ (m/s) | 0.059 | 0.006 | 8.892 | 0.000* | 1.062 |
| $L_{V_i}$ (m) | -0.008 | 0.048 | -0.162 | 0.872 | 0.992 |
| $V_{min}$ (m/s) | -0.038 | 0.010 | -3.811 | 0.000* | 0.962 |
| Average Deceleration Rate $d_m$ (m/s$^2$) | -0.245 | 0.016 | -15.343 | 0.000* | 0.782 |
| PCW system | 0.074 | 0.031 | 2.423 | 0.010* | 1.077 |
| Gender - Male | 0.420 | 0.144 | 2.914 | 0.004* | 1.522 |
| Not familiar with downtown Baltimore | 0.104 | 0.060 | 1.723 | 0.085 | 1.110 |
| Familiar with CAVs | -0.060 | 0.042 | -1.417 | 0.156 | 0.942 |
| Scale Parameter P | 2.628 | 0.085 | | | |
| AIC | -382.730 | | | | |
| Log-likelihood at convergence | -128.158 | | | | |
| Number of groups | 103 | | | | |

* Statistically significant at 99% CI

    Table 4 identifies the variables that are statistically significant to the participants' speed reduction times, in response to the jaywalking pedestrian. The variables significant at a 99% confidence interval were the initial speeds recorded at the waypoint, the minimum speeds reached in the deceleration phase, the average deceleration rates, the PCW system compared to the baseline, and gender. If the initial speed increases, the speed reduction time would also indirectly increase by 6.2% (odds ratio = 1.062), whereas the speed reduction times would decrease by 3.8% (odds ratio = 0.962) and 21.8% (odds ratio = 0.782) if there is a decrease in the minimum speed and the average deceleration rate. In the presence of a PCW system, the participants' speed reduction time increases by 7.7% (odds ratio = 1.077), which infers that the PCW system is the more effective system in improving speed reduction times, i.e., more time to transition to an acceptable speed or come to a stop, to yield to the pedestrian. Odds of male participants having a higher speed reduction time on average was 52.2% more than their female counterparts (odds ratio = 1.522), which implies that the male participants braked aggressively initially, and then proceeded slowly, until the pedestrian had passed. A scale parameter estimate of 2.628 implies that the survival probability of the speed reduction times decreases with the passage of time.

    A representation of the participants' braking patterns can be shown by plotting survival curves of the speed reduction times for the baseline scenario and the scenario involving the PCW system. These predictions were done based on the predict survival regression tool in the R-package [43] and shown in Figure 7.

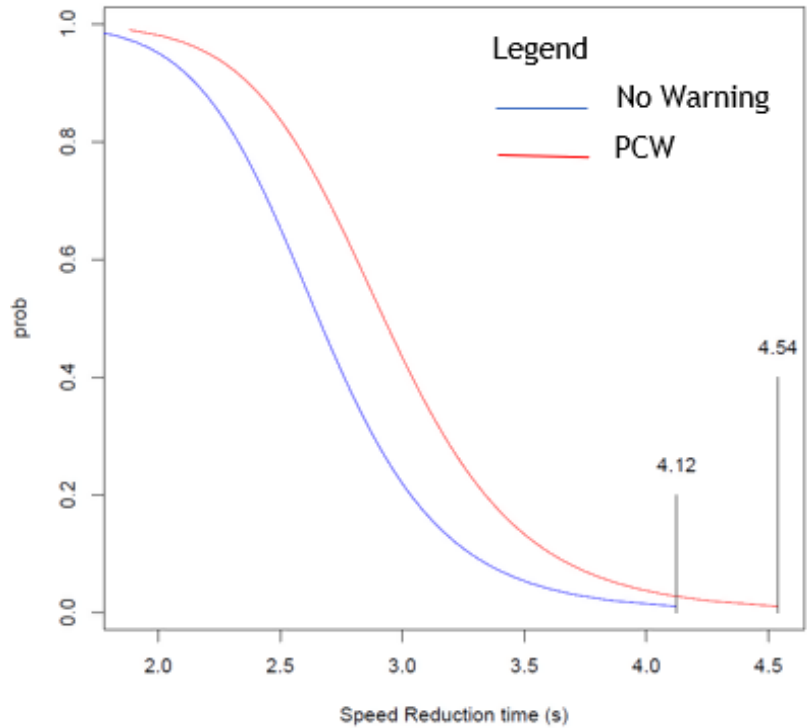
**Figure 7. Speed reduction time survival curves**

From Figure 7, it can be seen that the speed reduction time survival probability decreases with the passage of time. A lower survival probability was recorded for the baseline scenario as compared to the scenario with the PCW system. At 3 seconds of speed reduction time, the survival probability for the baseline scenario was 21% compared to 43% for the scenario with the PCW system, and it drops even further at 4.5 seconds, to only 5% in the baseline scenario to 13% in the PCW system scenario. The speed reduction time was 0.42 seconds longer (statistically significant) in the presence of a PCW system, giving the participants longer time to brake and transition to a safe stop.

*3.3. Eye Gaze Analysis*

This study used Tobii Pro Glasses 2 [17], the eye tracking device and its analysis software. The data was extracted and merged with the driving simulator using a novel application [44]. Both the scenarios, the baseline and the one with the PCW system, were evaluated, to analyze where the participants glanced while at the waypoint. If the participant glances at any object for a moment, it is captured by the eye tracking device. In this analysis, the glance is recorded at the waypoint, which is when the jaywalking pedestrian first becomes visible. A heat map can be generated using the number of fixations the participants made in certain areas, with red indicating the highest number of fixations and green indicating the least number of fixations. The generated heat map just shows a count of the number of fixations on the object or area of interest. An eye gaze heat map analysis of the 103 participants, the moment the vehicles cross the 40-meter threshold, is shown in Figure 8.

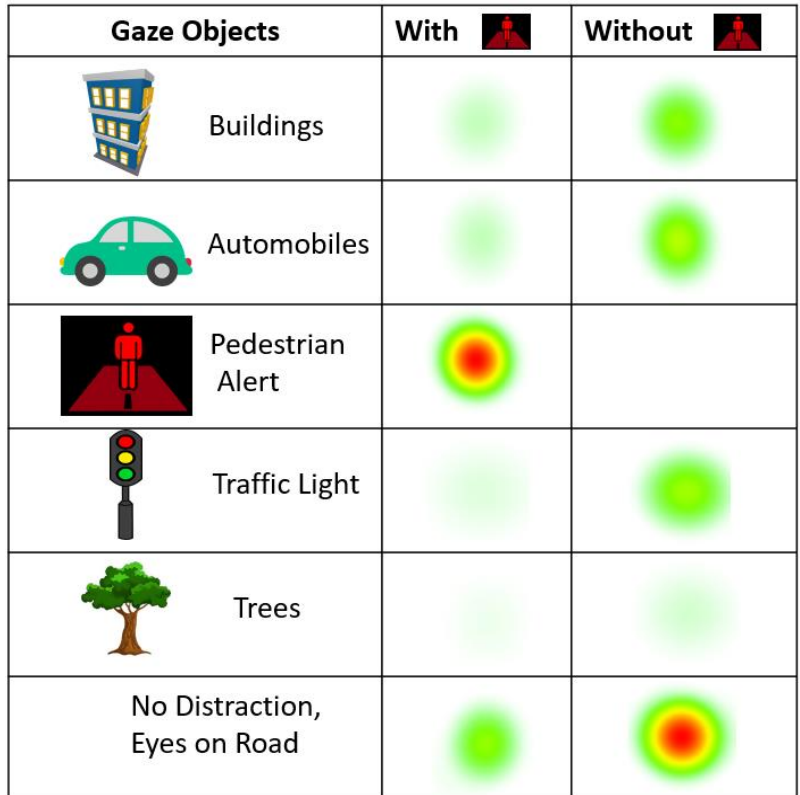

**Figure 8. Eye tracking gaze analysis**

Figure 8 clearly shows that the participants glanced at the PCW and this possibly affected their speed reduction times compared to the baseline scenario in which even though they may not be distracted, it may be difficult to observe the pedestrian at that very moment. This infers that having a PCW system informs the driver of an oncoming pedestrian, giving them plenty of time to react accordingly. The gaze data also shows that participants who glanced at the PCW popup had longer speed reduction times compared to participants who did not look at the popup.

From the videos recorded using the eye tracker, it was observed that, of the 27% of the participants who did not stop or ran over the jaywalking pedestrian in the baseline scenario, 64% of these participants did actually stop for the pedestrian in the PCW system scenario, confirming the benefits a PCW system. The authors did not find any significant correlation with sociodemographics of the participants and eye gaze analysis.

**4. Conclusions**

The application, a PCW system, was tested on driver braking behavior using a full-scale, medium fidelity driving simulator and an eye tracking device. Ninety-three participants were involved in this study, contributing to 186 simulation sessions. The scenarios were evaluated for the participants' braking behavior at the sudden appearance of a jaywalking pedestrian in the form of speed reduction times using a Log logistic AFT model. It was observed that speed reduction times were higher in the presence of a PCW system and men were more inclined to have a higher speed reduction time compared to women, leading to an initial aggressive braking behavior. This suggests that, the presence of a PCW system sends a clear message to the driver about the presence of a pedestrian, giving the driver ample time to adapt their initial approach speed to yield to the pedestrian. Although, the warning may force the drivers to brake aggressively initially, the objective of the system to protect the pedestrian will be fulfilled. These findings were substantiated by the jerk analysis, which shows that, jerk is unsafe, in the initial seconds of a PCW, while the scenario without the PCW system, results in a highly uncomfortable jerk, when the pedestrian is closer to the vehicle. These results confirm those obtained by [13] and [45] whose pedestrian warning systems had significant impacts on the drivers' performance; albeit the results from [46] contradict these findings as this study found no significant impact of the system on the deceleration rate. The overall analysis also suggested that high initial approach speeds, decrease in minimum speeds during the deceleration phase, and a decrease in average deceleration rates have a significant positive influence on the speed reduction times. An interesting observation from the post simulation survey questionnaire was that 31% of the participants said that a PCW system was their top choice of CAV technology to have in their automobiles. The eye tracking analysis showed that majority of the participants did glance at the PCW as compared to observing other objects; its absence may impact the reaction time needed to stop for the pedestrian. A longer speed reduction time means that, although this may involve aggressive braking initially, the drivers will eventually pace themselves at a speed that will allow the pedestrian to pass safely. This may also be true for drivers who exceed the safe speed limit of the road and may not brake appropriately to stop for the pedestrian in time, if not for the PCW. Although the PCW systems had longer speed reduction times, the participants braked aggressively initially, at the onset of the warning, before gradually slowing down and coming to a stop. This may not be the ideal scenario in terms of avoiding rear end collisions at the moment, but it will possibly help prevent pedestrian crashes. Consequently, it is anticipated that these systems will be more widely adopted by the automobile manufacturers and come prebuilt into the vehicles rather than an optional package, as they will help improve the safety of all road users and reduce congestion. As drivers become more accustomed to the technology, they may not need to brake aggressively initially and could be more gradual throughout the deceleration phase.

Although, predictive emergency braking systems is a relatively new technology [47], its influence on driver behavior needs to be researched. Future studies would involve combining PCW with automatic emergency braking systems and evaluating driver reaction times, in this scenario.

## 5. Acknowledgements

This study was supported by the Urban Mobility and Equity Center, a Tier 1 University Transportation Center of the U.S. DOT University Transportation Centers Program at Morgan State University. The authors would like to acknowledge the guidance and support received from Dr. Young-Jae Lee, Associate Professor at Morgan State University.